\documentclass[11pt,a4paper]{article}
\usepackage[hyperref]{eacl2021}
\usepackage{times}
\usepackage{latexsym}

\usepackage{url}
\usepackage{textcomp}
\usepackage{hyperref}
\usepackage{amsmath}
\usepackage{amssymb}
\usepackage{graphicx}
\usepackage{subfig}
\usepackage{multirow}
\usepackage{bm}

\usepackage{microtype}

\aclfinalcopy 

\title{Adaptive Fusion Techniques for Multimodal Data}

\author{Gaurav Sahu, Olga Vechtomova \\
    University of Waterloo \\
  \texttt{\{gsahu, ovechtom\}@uwaterloo.ca} \\}

\date{}

\begin{document}
\maketitle
\begin{abstract}
Effective fusion of data from multiple modalities, such as video, speech, and text, is challenging due to the heterogeneous nature of multimodal data.
In this paper, we propose adaptive fusion techniques that aim to model context from different modalities effectively.
Instead of defining a deterministic fusion operation, such as concatenation, for the network, we let the network decide ``how" to combine a given set of multimodal features more effectively.
We propose two networks: 1) Auto-Fusion, which learns to compress information from different modalities while preserving the context, and 2) GAN-Fusion, which regularizes the learned latent space given context from complementing modalities.
A quantitative evaluation on the tasks of multimodal machine translation and emotion recognition suggests that our lightweight, adaptive networks can better model context from other modalities than existing methods, many of which employ massive transformer-based networks.\footnote{Code for our experiments: \url{https://github.com/demfier/philo/}}
\end{abstract}

\section{Introduction}
Multimodal deep learning is an active field of research, where for a single event, one has information across multiple modalities, such as video, speech, and text.
Human brains can easily and perpetually perceive the context of an event from such heterogeneous data;
however, it is not a trivial task for a computer system.
In order for the machine to gain a contextual understanding, heterogeneous inputs must be combined first.
Combining, or more precisely, \textit{fusing} multimodal inputs is, thus, a vital step for any multimodal task.
Naturally, a better fusion method will help a multimodal system learn better, ultimately enhancing its performance for a given task.

The most common fusion technique used in the literature involves the concatenation of representations from all the available modalities.
However, this results in a shallow network \citep{ngiam2011multimodal}, and the network focuses more on learning intra-modal features, ignoring inter-modal dynamics altogether.
Later, \newcite{zadeh2017tensor} proposed Tensor Fusion Network (TFN), which models the unimodal, bimodal, and trimodal interactions using a 3-fold Cartesian product.
TFN performs better than simple concatenation;
however, it imposes high computational requirements since it projects all the information from input modalities to a dense 3-D space as-is, without any prior information extraction.
The computational overhead grows exponentially with respect to the dimensionality of unimodal features.
\newcite{liu2018efficientlow} proposed a low-rank multimodal fusion technique (LMF) to address the previous problem.
Such fusion techniques are useful but often result in a complex architecture.
Moreover, the fusion methods mentioned above focus only on combining individual unimodal features rather than combining \textit{and} extracting useful information simultaneously.
This means that the final predictor module (decoder in a Seq2Seq network \citep{sutskever2014sequence}, for example) bears an additional responsibility of identifying useful signals to focus on.

This paper addresses these issues by proposing adaptive fusion techniques that allow the model to decide ``how" to combine multimodal data more effectively for an event.
The first technique, Auto-Fusion, learns to compress multimodal information while preserving as much meaning as possible.
The second technique, GAN-Fusion, employs an adversarial network that regularizes the learned latent space for a given target modality complying with the information presented by complementary modalities.
Since our models are generic, the need to specify a pre-determined fusion operation such as Cartesian product is alleviated, and this further incentivizes the network to model multimodal interactions by itself.
Moreover, our techniques are lightweight relative to the existing heavier counterparts \citep{vaswani2017attention, gronroos2018memad}, thereby preventing unnecessary computational load.

We evaluate our models on three benchmark datasets:

\begin{enumerate}
    \item the How2 dataset \citep{sanabria2018how2} with multimodal input for English-Portuguese translation.
    \item the Multi30K dataset \citep{elliott-EtAl:2016:VL16}, which contains parallel corpora for multimodal machine translation, and
    \item the IEMOCAP dataset \citep{busso2008iemocap} which contains multimodal data for emotion detection.
\end{enumerate}

A quantitative evaluation shows that our models outperform the existing state-of-the-art methods in terms of BLEU scores \citep{Papineni2002bleu} for machine translation and Precision, Recall, and F1-score for emotion recognition.
Our ablation studies also indicate that the learned multimodal representations are robust;
they perform reasonably well even after removing information from a target modality.
We now summarize our main contributions as follows:

\begin{enumerate}
    \item We propose two lightweight, adaptive techniques for better multimodal fusion of data: Auto-Fusion and GAN-Fusion.
    \item We propose a multi-task framework for end-to-end training of multimodal networks (for both classification and generation).
\end{enumerate}

The rest of the paper is structured as follows: Section \ref{sec:rel} covers relevant work, Section \ref{sec:prop_meth} discusses the proposed methodologies and overall architecture, Section \ref{sec:exp} describes the experimental setup, Section \ref{sec:result} shows results, and Section \ref{sec:cncl} contains our concluding remarks.

\begin{figure*}[t]
	\centering
	\subfloat[Auto-Fusion network]{\includegraphics[scale=0.27,clip]{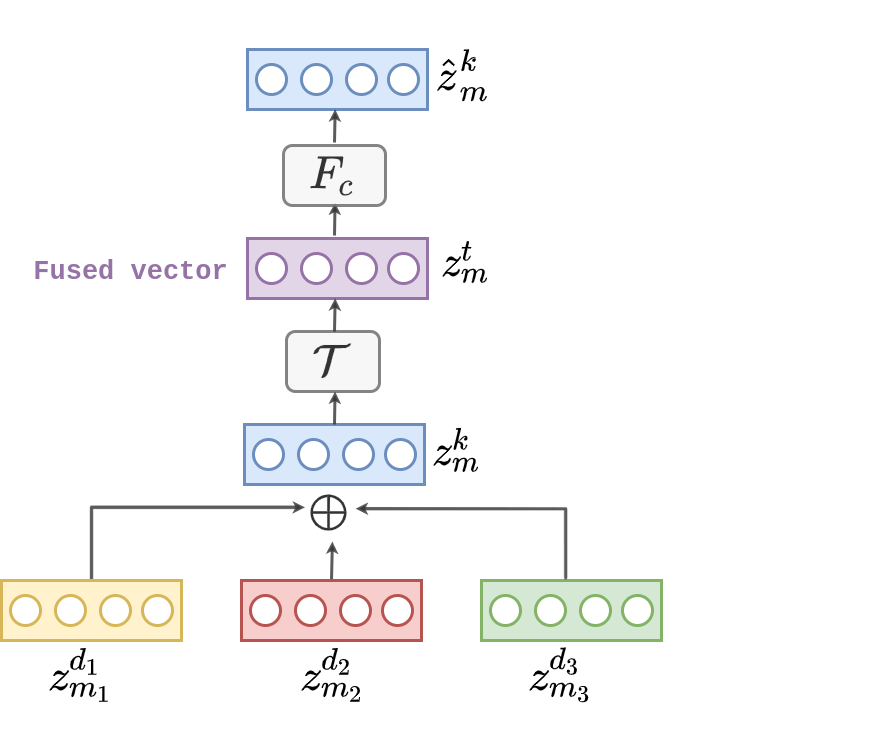}}
	\subfloat[GAN-Fusion Network]{\includegraphics[scale=0.3,clip]{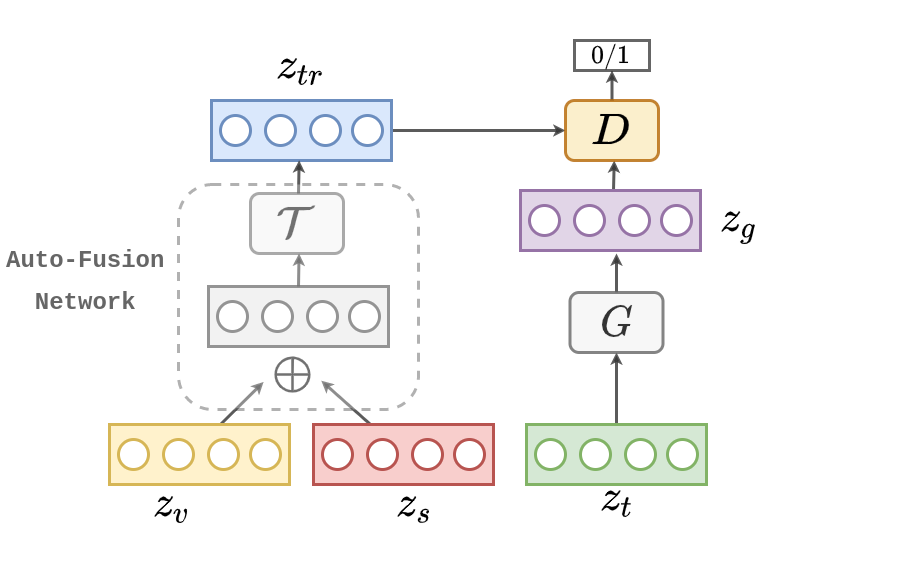}}
	\caption{Proposed architectures. (a) Auto-Fusion network: Assuming that $\bm {z_{m_1}^{d_1}}$, $\bm {z_{m_2}^{d_2}}$, and $\bm {z_{m_3}^{d_3}}$ represent the video, speech, and text latent vectors respectively, we first concatenate them to obtain $\bm {z_{m}^k}$. It is then passed through $\mathcal{T}$ which outputs the ``autofused" vector $\bm {z_{m}^t}$. We then obtain the reconstructed concatenated vector $\bm \hat z_{m}^k$ by passing the autofused vector through $F_c$, another transformation layer. Finally, we optimize the loss between $\bm {\hat z_{m}^k}$ and $\bm z{_{m}^k}$. (b) GAN-Fusion module for the text modality: Assuming that $\bm {z_s}$, $\bm {z_v}$, and $\bm {z_t}$ are the latent speech, video, and text vectors, respectively, we first autofuse $\bm z{_s}$ and $\bm {z_v}$ to give $\bm {z_{tr}}$. Simultaneously, we pass $\bm {z_t}$ through the generator $G$, along with some noise, to get $\bm {z_g}$. The generator loss tries to match $\bm {z_{tr}}$ and $\bm {z_g}$ and discriminator $D$ tries to distinguish between $\bm {z_{tr}}$ and $\bm {z_g}$, the two sources of input.
		\textbf{Note:}  $\bigoplus$ denotes concatenation.}
	\label{fig:models}
\end{figure*}

\section{Related Work}
\label{sec:rel}

In this section, we briefly review previous work related to our task. 
Most earlier works in multimodal deep learning focus on traditional shallow classifiers such as support vector machines \citep{cortes1995support} and Naive Bayes classifiers \citep{Morade2015} to exploit bimodal data.
Inspired by the success of deep learning over the last decade across multiple tasks, \citet{ngiam2011multimodal} train end-to-end deep graph neural networks to reconstruct missing modalities during inference.
They demonstrate that better features for one modality can be learned if relevant data from different modalities is available at training time; however, they employ simple concatenation for fusion.
Hence, the joint representation learned is shallow and is not guaranteed to model inter-modal interactions.
Their findings were later verified by \newcite{srivastava2012multimodal}, who use a Deep Boltzmann Machine \citep{salakhutdinov2009deep} to generate data from the image and text modality. \citet{huang2018multi} construct a multilingual common semantic space to achieve better machine translation performance by extending correlation networks \citep{chandar2016correlational}.
They use multiple non-linear transformations to reconstruct sentences from one language to another repeatedly and finally build a common semantic space for all the different languages.
To address the shallowness exhibited by some earlier fusion methods, techniques such as TFN \citep{zadeh2017tensor}, LMF \citep{liu2018efficientlow} and T2FN \citep{liang2019learning}, were proposed that aim to capture \textit{both} intra- and inter-modal dynamics simultaneously; however, the problem of effectively modelling context in multimodal samples remains unsolved.

More recently, Multimodal Transformer (MulT) \citep{tsai2019multimodal} was proposed to align data from different modalities implicitly.
On a high-level, MulT leverages cross-modal attention modules for each modality, each of which is responsible for aligning (or attending to) the target modality vector with the complementary modalities'.
It also imposes substantial computational overhead due to the use of transformer networks \citep{vaswani2017attention}.
Our methods, as discussed in detail in Section \ref{sec:prop_meth}, use much simpler components.
For instance, we use at most one attention module compared to multiple self-attention heads in a transformer.
Variational Mixture-of-Experts Autoencoders \citep{shi2019variational}, a class of deep generative multimodal frameworks, were employed to learn a synergic shared representation for multiple modalities;
however, scaling of such a model for all the modalities (video, speech, and text) simultaneously and for a more complex task as multimodal machine translation is currently unexplored.

\section{Proposed methods}
\label{sec:prop_meth}
This section will discuss the proposed methodologies for effectively fusing inputs from multiple modalities and describe the overall architecture of our models for classification and generation.
Most fusion techniques proposed in the literature, such as concatenation, and TFN, involve a deterministic operation for constructing the joint multimodal representation.
For instance, in TFN, the 3-fold Cartesian product of unimodal features is used for prediction.
The method focuses more on learning rich unimodal features.
However, there is no such ``learning" procedure for joint representation;
they are simply constructed by combining unimodal features in a specific fashion (here, by Cartesian product.)
In this paper, we will refer to such techniques as \textit{static} fusion techniques.
Since there is no particular learning procedure for the joint representation, it becomes challenging for the final predictor module to model the complex dynamics of multimodal features.
In other words, the model is unable to utilize multimodal information effectively.

On the other hand, fusion methods such as LMF and MulT are \textit{adaptive} because they involve a cognitive feature processing step to construct the joint representation.
In LMF, it is the decomposition module, and in MulT, it is the final feed-forward fusion mechanism.
We refer the reader to \newcite{liu2018efficientlow} and \newcite{tsai2019multimodal} for more detailed explanation of the models.

Our fusion methods involve  the concatenation of unimodal embeddings as an initial step.
To avoid any conflicts with past works, we will only consider steps \textit{after} concatenation as a part of our fusion method because we do not use the concatenated vector for final prediction; it is only a preliminary step.
Therefore, in order to mitigate the ``staticness" of existing fusion methods, we propose two adaptive yet simple fusion techniques, \textit{Auto-Fusion} and \textit{GAN-Fusion}.
They aim to effectively combine multimodal inputs and mitigate the problem of shallowness and computational overhead exhibited by prior fusion techniques.

\begin{figure}[t]
    \centering
    {\includegraphics[scale=0.31,clip]{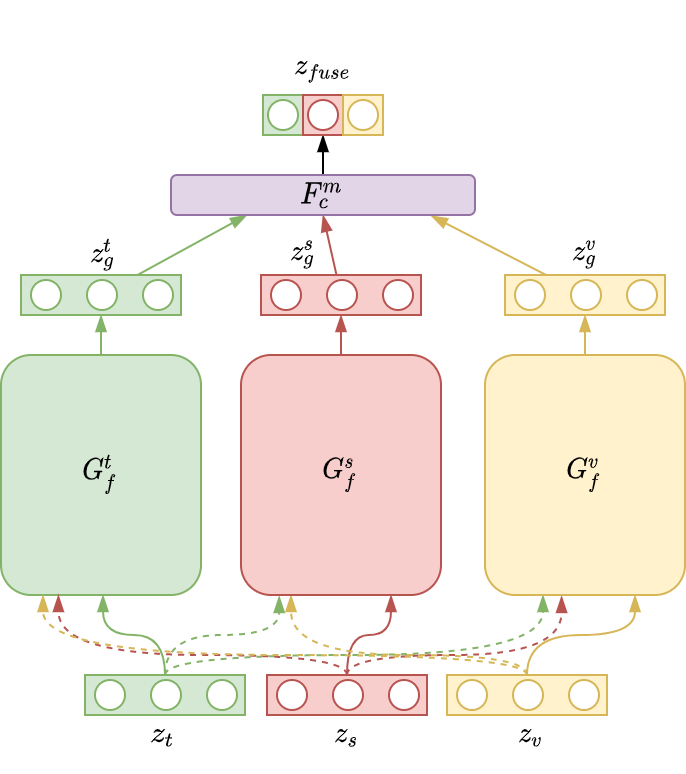}}
    \caption{Overall architecture of the GAN-Fusion module. Here, the solid and dashed lines at the bottom part represent input from the target and the complementary modalities, respectively, for $G_{f}^{m}$, the GAN-Fusion module with target modality $m \in \{t, s, v\}$. Furthermore, $F_{c}^{m}$ represents the feed-forward layer that produces the fused multimodal representation $\bm{z_{fuse}}$, which is, in turn, fed to the target decoder for generation networks and to the fully-connected network for classification networks.}
    \label{fig:gan_fusion_complete}
\end{figure}

\subsection{Auto-Fusion}
This method encourages the model to extract intermodal features by maximizing the correlation between multimodal inputs.
In this method, we first concatenate individual unimodal features and then pass them through a transformation layer to get a \textit{autofused} latent vector.
We use appropriate learners for individual modalities (See Section \ref{sec:exp}).
We then try to reconstruct the originally concatenated vector from the autofused latent vector.
Finally, we minimize the Euclidean distance between the original and reconstructed concatenated vector.
This process ensures that the learned autofused vector does not contain arbitrary signals from the input concatenated latent vector.
Furthermore, training the model for a downstream task such as emotion recognition incentivizes it to ``compress" information without losing any essential cues.
In other words, it increases the correlation between the autofused and the concatenated latent vector.
This generic procedure applies to any scenario where multiple features need to be combined.
For example, it can even be used to combine the forward and backward hidden states of LSTMs \citep{hochreiter1997long}, instead of pooling methods such as 1D pooling, max pooling, sum pooling or even simple concatenation.

We now discuss the Auto-Fusion network in detail.
We pose fusion of multimodal inputs as a compression problem, where we must retain as much information from the individual modalities as possible.
Given $n$ ($\leq3$ in our case) $d$-dimensional multimodal latent vectors, $\bm z_{m_1}^{d_1}, \bm z_{m_2}^{d_2},\dots,\bm z_{m_n}^{d_n}$, we first concatenate them to obtain a vector, $\bm z_{m}^k$, where $k = \sum_i^n d_i $.
Then, we apply a transformation $\mathcal{T}$ to $\bm{z_{m}^k}$, reducing its number of dimensions to $t$.
Then, we use $\bm{z_{m}^t}$ to reconstruct the originally concatenated vector $\bm{\hat z_m^{k}}$.
Finally, we calculate the loss, $J_{tr}$, between $\bm{\hat z_m^{k}}$, and $\bm{z_{m}^k}$.
The simplest version of Auto-Fusion network employs the mean squared error (MSE) loss function, which aligns with our motivation to compress multimodal features: filter out the less useful signals.
These steps could be followed in Figure \ref{fig:models}(a) and the MSE loss for Auto-Fusion network is given by:

\begin{equation}
J_{tr} = || \; \bm{\hat z_{m}^k} - \bm{z_{m}^k} \; ||^2
\label{eq:autoloss}
\end{equation}

For Auto-Fusion, we consider the intermediate vector, $\bm{z_m^{t}}$, as the fused multimodal representation.

\subsection{GAN-Fusion}
In addition to the ``staticness" of existing methods, there is also the challenge of distinguishing between ambiguous cases.
For instance, the sentence ``Your joke blew my mind away, Kevin," could be said in a funny or sarcastic manner.
Resolving ambiguity becomes especially important when working on social problems such as hate speech detection.
Even when fed with the corresponding speech vector, existing methods cannot effectively distinguish between similar but different emotions such as happiness and calmness.
We hypothesize that this is because they do not learn the conditional distribution of sentiment given an utterance (an utterance includes input from all available modalities).

To address this issue, we propose an adversarial training regime that is incentivized to learn the desired conditional distribution.
For a task such as emotion recognition, the objective would be sentiment given an utterance.
For a more challenging generation task, the model could learn a more complex behaviour, such as the association of different sentences based on how similar they sound and their polarity.
Our experiments show that our GAN-based approach is better able to learn such multimodal dynamics compared to other methods.


We now describe GAN-Fusion's architecture in detail for target modality text (denoted by $t$.)
For a given multimodal sample $x$, we first encode the inputs from each modality (speech, visual and text) to get the respective latent vectors, $\bm{z_s}, \bm {z_v}, \textrm{ and } \bm {z_t}$.T
Choosing a target modality such as text, we pass $\bm {z_t}$ (along with random normal noise,) through a generator to obtain $\bm z_g = G(\bm {z_t})$ and autofuse the remaining latent vectors $\bm {z_s} \textrm{ and } \bm {z_v}$ simultaneously to obtain $\bm {z_{tr}}$.
In the event where we have input from only one modality in addition to text, we do not need Auto-Fusion, and can simply treat the other modality's vector as $\bm{z_{tr}}$.
Finally, we train the network in an adversarial fashion, labelling $\bm{z_{tr}}$ as positive samples and $\bm{z_g}$ as negative samples.
The adversarial loss, $J_{adv}^{t}$, is given below:
\begin{equation}
    \begin{split}
        \min_G \max_D J_{adv}^{t} (D, G) &= \mathbb{E}_{x \sim p_{\bm {z_{tr}}}(x)}[\textrm{log} D(x)] \\
        &+ \mathbb{E}_{\bm z \sim p_{z_t} (\bm z)}[\textrm{log}(1-D(\bm z_g))]
    \end{split}
    \label{eq:ganloss}
\end{equation}

Overall, the generator $G$ tries to align features of the target modality with features from the complementary modalities, and the discriminator tries to discern the source of its input.
Such a translation between latent vectors has been shown to learn an ``intermediate" latent vector denoting their joint representation \citep{pham2019found, gao2019jointly}.
Learning the latent space in such an adversarial manner induces a clustering effect on the latent space, where texts associated with similar sounds and visuals are grouped together.
We conjecture that the adversarial training helps the model learn the relative topology of the complementary modalities' latent space, which improves sampling from the target modality.

\begin{figure}[t]
	\centering
	\includegraphics[scale=0.3,clip]{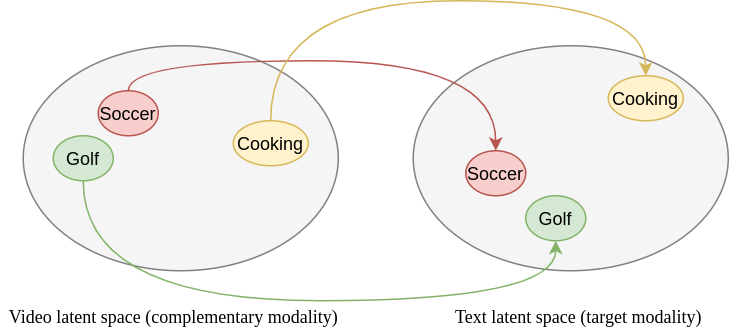}
	\caption{Visualizing the induced clustering effect of GAN-Fusion.
	The circles represent a cluster of words related to the indicated topic (text inside the circles)}
	\label{fig:gan_clustering}
\end{figure}

We elucidate this effect through the following example.
For the sake of simplicity, let us consider only one complimentary modality (video) in this case and let text be the target modality.
First, we make a reasonable assumption that video embeddings for, say, Soccer and Golf--falling under the general category of Sports--will be mapped closer to each other and farther from video embeddings from an unrelated topic such as Cooking.
When learning the intermediate representation between video and text, the text latent space is constructed such that its relative topology partially reflects the video latent space.
So, token embeddings in the text latent space related to videos of similar events (Soccer and Golf) will adopt similar relative positioning as followed by the video embeddings for Soccer and Golf in the video latent space.
For multimodal machine translation, if the model is fed with a Golf video and the source text as input, it may be better able to sample jargon words for Golf from the text latent space due to this topology inheritance.
This ultimately improves translation-quality.
This is also depicted in Figure \ref{fig:gan_clustering}.

\begin{figure}[t]
	\centering
	\includegraphics[scale=0.26,trim={16mm 5mm 0 0},clip]{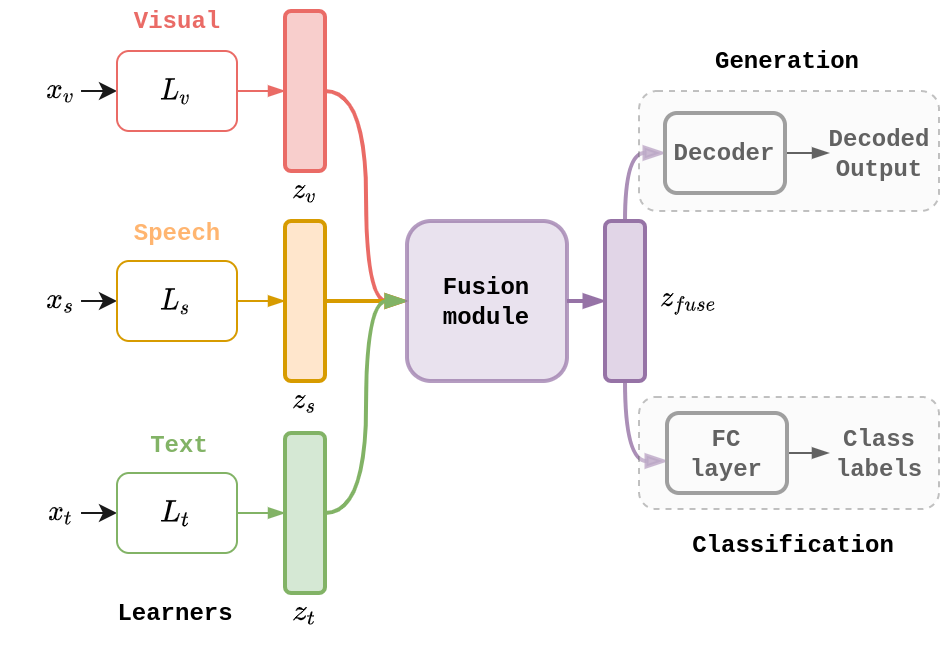}
	\caption{Using proposed fusion techniques for generation/classification. Unimodal inputs $x_v, x_s, x_t$ are passed through their respective learners $L_v, L_s, L_t$ to obtain unimodal representations $\bm{z_v}, \bm{z_s}, \bm{z_t}$. Here, $v, s, t$ correspond to visual, speech, and textual modalities respectively. The individual unimodal representations are then passed through the fusion module (either Auto-Fusion or GAN-Fusion,) which outputs the fused multimodal representation $\bm{z_{fuse}}$. For generation, $\bm{z_{fuse}}$ is then passed through a compatible decoder, which generates outputs for the desired target modality. For classification, $\bm{z_{fuse}}$ is passed through a fully-connected layer instead, which predicts appropriate the class labels.}
	\label{fig:overall_net}
\end{figure}

Figure \ref{fig:models}(b) shows GAN-Fusion module for the text modality.
The GAN-Fusion module, overall, has one such module for every modality.
Total adversarial loss is, therefore, given by:

\begin{equation}
    J_{adv} = J_{adv}^t + J_{adv}^s + J_{adv}^v
    \label{eq:ganloss_full}
\end{equation}

where losses $J_{adv}^{s}$ and $J_{adv}^{v}$ for speech and video, respectively, are defined similarly as $J_{adv}^{t}$.

Figure \ref{fig:gan_fusion_complete} shows an overall architecture of the GAN-Fusion module, which consists of $G_{f}^{t}$, $G_{f}^{s}$, and $G_{f}^{v}$, the respective fusion modules for text, speech, and video modalities.
We pass the outputs of these modules through a feed-forward layer to obtain the final fused multimodal representation $\bm{z_{fuse}}$.

\subsection{Overall Architecture}
In this section, we describe the end-to-end training process for using the proposed fusion methods for 1) Generation tasks (e.g. visual question answering, multimodal machine translation) and 2) Classification tasks (e.g. speech emotion recognition, hate speech detection.)

\textbf{Generation:} Figure \ref{fig:overall_net} shows the end-to-end pipeline integrating proposed fusion techniques for generation.
We first pass raw inputs from different modalities through their respective learners to obtain their respective latent representations.
They are then passed through the fusion module that outputs a fused representation to be used for decoding.
We only validate this process for generating text, but this method could very well be used for generating outputs for different target modalities.
Notably, our pipeline for generation looks very similar to a Seq2Seq network.
We simply introduce a fusion module between the encoder and the decoder module.
It should also be noted that all GAN-Fusion modules for all the target modalities ($G_{f}^{t}, G_{f}^{s}, G_{f}^{v}$) are trained simultaneously with the rest of the network.

\textbf{Classification:} Figure \ref{fig:overall_net} also shows how to adapt the previously described generation network for classification: simply replace the decoder with a fully-connected layer to predict appropriate class labels.

It is important to note that $\bm{z_{fuse}}=\bm{z_{m}^{t}}$ for Auto-Fusion, and it is obtained as shown in Figure \ref{fig:gan_fusion_complete} for GAN-Fusion.
The overall loss function for our networks can be generalized as follows:

\begin{equation}
J_{total} = \lambda_1 J_{fusion} + \lambda_2 J_{task}
\end{equation}

Here, $J_{fusion}$ refers to the loss function of the fusion network.
It equals $J_{tr}$ (from equation \ref{eq:autoloss}) when using Auto-Fusion, and $J_{adv}$ (from equation \ref{eq:ganloss_full}) when using GAN-Fusion.
Furthermore, $J_{task}$ refers to the task-specific loss, i.e., classification loss (such as max-margin loss) or generation loss (such as cross-entropy loss for Seq2Seq network).
$\lambda_1 \text{ and } \lambda_2$ are hyperparameters to tune.

\section{Experimental Setup}
\label{sec:exp}

We measure our models' effectiveness on two tasks: 1) multimodal machine translation and 2) multimodal emotion recognition.
The subsequent sections describe our complete experimental setup, including datasets and baselines used, implementation details, and evaluation metrics.

\subsection{Datasets}
We choose three datasets for our experiments, which are described as follows:

\begin{table*}[ht]
    \small
    \centering
    \begin{tabular}{|c|c|c|c|c|c|}
        \hline
        \textbf{Model} & \textbf{Source modalities} & \textbf{BLEU 1} & \textbf{BLEU 2} & \textbf{BLEU 3} & \textbf{BLEU 4} \\
        \hline
        Unimodal S2S & t & - & - & - & 54.4 \\
        \hline
        Multimodal S2S & s-v-t & - & - & - & 54.4 \\
        \hline
        BPE Multimodal & s-v-t & - & - & - & 51.0 \\
        \hline
        Unimodal SPM Transformer & t & - & - & - & 55.5 \\
        \hline
        Attention over Image Features & s-v-t & - & - & - & 56.2 \\
        \hline
        \multirow{3}{*}{\textbf{Seq2Seq (w/o attn)}} & t & 48.32 & 30.63 & 20.79 & 14.60 \\
                                 \cline{2-6}
                                 & s & 20.11 & 7.01 & 3.12 & 1.57 \\
                                 \cline{2-6}
                                 & v & 19.28 & 6.35 & 2.33 & 1.03 \\
        \hline
        \textbf{Seq2Seq} & t & 79.21 & 67.34 & 52.67 & 47.34 \\
        \hline
        \multirow{2}{*}{\textbf{Auto-Fusion (Ours)}} & s-t & 80.34 & 67.83 & 61.27 & 55.01 \\
                                                \cline{2-6}
                                                & s-v-t & 85.23 & 71.95 & 69.54 & 57.80 \\
        \hline
        \multirow{2}{*}{\textbf{GAN-Fusion (Ours)}} & s-t & 82.25 & 69.43 & 64.33 & 56.5 \\
                                                \cline{2-6}
                                                & s-v-t & \textbf{89.66} & \textbf{74.48} &\textbf{ 71.29} & \textbf{59.83} \\
        \hline
    \end{tabular}
    \caption{Results for machine translation on How2 dataset. `t', `s', `v' represent the text, speech, and video modalities, respectively. Here, `attn' refers to the word-level attention \citep{luong2015effective}.}
    \label{tab:mt_how2}
\end{table*}

\begin{table}[ht]
    \small
    \centering
    \begin{tabular}{|c|c|c|}
        \hline
        \textbf{Model} & \textbf{BLEU 4} & \textbf{Meteor} \\
        \hline
        Unimodal Seq2Seq & 36.3 & 56.9 \\
        \hline
        MeMAD submission & 44.1 & \textbf{64.3} \\
        \hline
        \textbf{Auto-Fusion (Ours)} & 42.31 & 61.7 \\
        \hline
        \textbf{GAN-Fusion (Ours)} & \textbf{44.23} & 63.8 \\
        \hline
    \end{tabular}
    \caption{Results for machine translation on Multi30K dataset. \textbf{Note:} All methods use `v' and `t' as the source modalities except Unimodal Seq2Seq.}
    \label{tab:mt_multi30k}
\end{table}

\begin{table}[ht]
    \small
    \centering
    \begin{tabular}{|c|c|c|c|c|}
        \hline
         \textbf{Model} & \textbf{P} & \textbf{R} & \textbf{F} & \textbf{A} \\
         \hline
         \textbf{LSTM (t)} & 53.2 & 40.6 & 43.4 & 43.6 \\
         \hline
         \textbf{LSTM ([s;t])} & 66.1 & 65.0 & 64.7 & 64.2 \\
         \hline
         MDRE & - & - & - & 71.8 \\
         \hline
         MHA-2 & - & - & - & 76.5 \\
         \hline
         \textbf{Auto-Fusion (Ours)} & 75.3 & 77.4 & 76.3 & 77.8 \\
         \hline
         \textbf{GAN-Fusion (Ours)} & \textbf{77.3} & \textbf{79.1} & \textbf{78.2} & \textbf{79.2} \\
         \hline
    \end{tabular}
    \caption{Precision (P), Recall (R), F1-score (F), and Accuracy (A) for emotion recognition on IEMOCAP.}
    \label{tab:emorec_iemocap}
\end{table}

\textbf{IEMOCAP:} We use the benchmark Interactive Emotional Dyadic Motion Capture (IEMOCAP) dataset \citep{busso2008iemocap} for emotion recognition.
We only use the textual and speech modalities for our emotion recognition experiments.
The dataset is originally split into multiple utterances for each session, and we further split each utterance file based on the provided start and end timestamps to obtain wav files for each sentence.
This results in a total of $\sim$10K audio files, which are then used to extract features for predicting a given utterance's emotion.
Concretely, we identify the task as an emotion recognition problem, where, given a sentence and its audio, we aim to infer the correct emotion for that utterance.

\textbf{How2:} We evaluate our models on the multimodal How2 dataset \citep{sanabria2018how2}, which comprises of 79,114 instructional videos, their Kaldi \citep{povey2011kaldi} audio features, and word-level time alignments of English-to-Portuguese translations.
The How2 dataset is trimodal in comparison to other multimodal datasets \citep{avletters, patterson2002cuave}.
This makes it suitable to evaluate the contribution of each modality for different tasks.
Further, as a large-scale multilingual dataset, it enables a convenient medium for neural machine translation in our work.

\begin{figure}[ht]
\centering
    \includegraphics[scale=0.42,trim={4mm 4mm 9mm 4mm},clip]{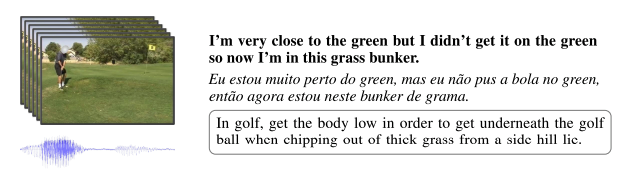}
    \caption{A multimodal sample from the \texttt{How2} dataset \citep{sanabria2018how2}}
    \label{figure:how2_example}
\end{figure}

\textbf{Multi30K:} In addition to the How2 dataset, we also run experiments on the bimodal Multi30K dataset \citep{elliott-EtAl:2016:VL16}--a benchmark dataset for machine translation--extended for German, where each sample has an image, its description in the source language, and its translated version.
We only run experiments on the En-Fr version, however.

\subsection{Implementation details}

\textbf{Generation:}
We train the network shown in Figure \ref{fig:overall_net} with Auto-Fusion and GAN-Fusion as fusion modules.
We use an LSTM encoder with 256 hidden units as the learner for textual description in our generation experiments, unless otherwise stated.
For the How2 dataset, we use the already provided 2048-dimensional feature vectors for video as raw input for the video learner, and we feed the Kaldi speech vectors to the speech learner, which is a simple feed-forward layer in this case.
For the Multi30K dataset, we use a pre-trained VGG \citep{simonyan2014very} to encode images, and we do not have a speech learner as there is no speech input in the dataset.
The latent dimension is 100 for the fused vector.

\textbf{Classification:}
For the task of speech emotion recognition, we train a multimodal classifier on the IEMOCAP dataset.
We use LSTMs with 256 hidden units to encode text.
For audio, we first pre-process the raw audio files to obtain a lower-dimensional feature vector and then use LSTMs with 50 hidden units as a learner.
We predict emotion labels through a full-connected layer as shown in Figure \ref{fig:overall_net} .
The latent dimension for the fused vectors is 50 in this case.

When training the GAN-Fusion network, we followed numerous tricks \citep{goodfellow2016nips} to ensure training stability.
A few that helped the most include input normalization, batch normalization, Leaky ReLU activation function \citep{Maas13} and Adam optimizer \citep{kingma2014adam} for the generator and discriminator networks.
The presence of multiple auxiliary losses in our networks also helped.

All the networks in our experiments are implemented in PyTorch \citep{pytorch}.
To train the different classification networks and generation networks on the Multi30K dataset, we use an Nvidia RTX 2080Ti with 12GB of RAM.
However, to train our trimodal networks on the How2 dataset, we use Nvidia P100 with 16 GB RAM to fit video feature vectors in memory.

\subsection{Baselines}
\textbf{How2:} We use the following baselines for experiments on the How2 dataset:

\begin{itemize}
    \item \textbf{Seq2Seq}: A sequence-to-sequence with attention mechanism \citep{luong2015effective}. It employs the previously described learners for each modality, and early fusion.
    \item \textbf{Unimodal and Multimodal S2S:} Unimodal and Multimodal S2S model as described in \newcite{sanabria2018how2}.
    \item \textbf{How2-challenge baselines:} We use Attention over Image Features (Wu et al., 2019), Unimodal SPM Transformer (Raunak et al., 2019) and BPE Multimodal (Lal et al., 2019) from the How2-challenge leaderboard.\footnote{\url{https://srvk.github.io/how2-challenge/}}
\end{itemize}

\textbf{Multi30K:} We use the following baselines for the Multi30K dataset:

\begin{itemize}
    \item \textbf{Unimodal Seq2Seq:} A text-only NMT system used by \newcite{ElliottFrankBarraultBougaresSpecia2017}.
    \item \textbf{MeMAD submission} \citep{gronroos2018memad}: The best performing model on Multi30K, a multimodal transformer network.
\end{itemize}

\textbf{IEMOCAP:} We use the following baselines for the IEMOCAP dataset:

\begin{itemize}
    \item \textbf{LSTM (t):} A unimodal LSTM classifier with attention mechanism trained using only text.
    \item \textbf{LSTM ([s;t]):} A bimodal LSTM classifier with attention mechanism on text only.
    We use the concatenation of speech and text features as the joint multimodal representation.
    \item \textbf{MDRE:} Multimodal Dual Recurrent Encoder proposed by \newcite{yoon2018multimodal}.
    \item \textbf{MHA-2:} A multimodal classifier with Multi-hop attention mechanism \citep{yoon2019speech}.
\end{itemize}

We compare the above baselines' performance with our two main models: GAN-Fusion and Auto-Fusion, which replace early fusion in the Seq2Seq baseline.
We report the results in Table \ref{tab:mt_how2}, \ref{tab:mt_multi30k}, and \ref{tab:emorec_iemocap}.

\subsection{Evaluation metrics}
We use Precision, Recall, F-Score, and classification accuracy to evaluate our classification networks trained for speech emotion recognition.
For experiments on the How2 and Multi30K dataset, we use BLEU \citep{Papineni2002bleu} to evaluate the quality of translated sentences.
For How2, we compute BLEU1-BLEU4 scores of the different models under consideration.
For Multi30K, we also use METEOR \citep{banerjee2005meteor}, which is a weighted harmonic mean of unigram precision and unigram recall providing a better indication of translation quality.
For all the mentioned evaluation metrics, a higher number denotes better performance.

\begin{figure}[t]
    \centering
    {\includegraphics[scale=0.4]{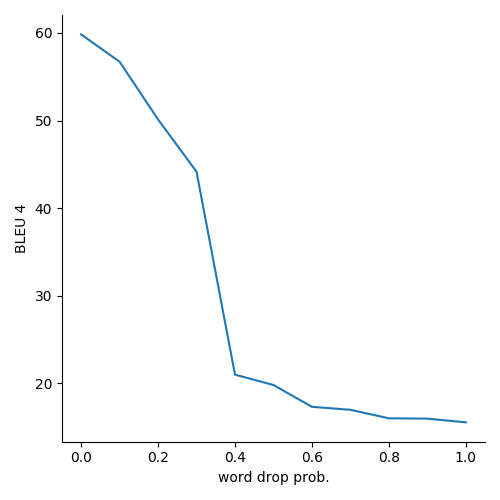}}
    \caption{Ablation test on the How2 dataset. Word drop probability v/s BLEU 4. A sudden drop in BLEU scores as we move from 0.3 to 0.4 indicates that our model was able to compensate for $\sim$ 30\% of the missing text.}
    \label{fig:ablation}
\end{figure}

\section{Results}
\label{sec:result}
\textbf{Quantitative analysis:}
Results of our experiments on the How2, Multi30K and IEMOCAP dataset are shown in Tables \ref{tab:mt_how2}, \ref{tab:mt_multi30k} and \ref{tab:emorec_iemocap}, respectively.
For speech emotion recognition, we observe that our models consistently perform well across all the evaluation metrics.
For the relatively challenging multimodal machine translation task, we observe that our model outperforms all existing baselines in terms of BLEU scores.
Compared against the best performing baselines, GAN-Fusion improves BLEU4 scores by 3.63 points and 0.13 points on the How2 and Multi30K datasets, respectively.
This shows that the fusion module was better able to extract signals from all modalities.
On the Multi30K dataset, our model is competitive in terms of METEOR.
Such performance becomes more pronounced considering our models have only one attention module, in contrast to multiple self-attention heads in the transformer-based baselines.

The size of our best-performing models (employing GAN-Fusion) is roughly 6M and 11M for classification and generation, respectively, which is significantly lower than traditional transformer models.
Fewer trainable parameters reduces the computational cost and the training time.
This also indicates potential by-passing of a mechanism like distillation, which is used to reduce parameters in transformers \citep{sanh2019distilbert};
however, more thorough experimentation is required to reach a concrete conclusion.
Our models can also be used in conjunction with transformers, where we utilize the transformers to learn meaningful unimodal feature vectors initially and then employ the proposed fusion methods.

We also perform a comprehensive set of qualitative experiments on the How2 dataset to understand the capability of our fusion techniques.
They are described as follows:

\textbf{Effect of introducing more modalities:}
To understand the effect of introducing new modalities separately, we perform experiments with different combinations of source modalities, including individual unimodal baselines (Refer to Table \ref{tab:mt_how2}.)
The results reveal that both auditory and visual modalities \textit{always} contribute towards enhanced translation, but the contribution of visual modality is slightly lower (indicated by the lower increase in BLEU scores in Table \ref{tab:mt_how2}.)
This is also consistent with the findings of \newcite{gronroos2018memad}.

\textbf{Robustness of multimodal features:}
It is very important for the learned multimodal latent features to be robust, i.e., they should be able to exploit signals from complementary modalities to compensate for the presence of noise in one modality.
Therefore, to gauge the robustness of learned multimodal features, we conduct an ablation test on the How2 dataset.
We randomly replace some tokens in the test sentence with an \texttt{<UNK>} token and attempt to translate using our best performing model, GAN-Fusion.
Figure \ref{fig:ablation} shows results of our ablation study.
We can see that features from the complementary modalities are able to compensate for $\sim30\%$ of the missing text as we see a sharp drop in BLEU scores beyond that point.
This shows that the model does not rely on just the textual description for translation; it also tries to gain a contextual understanding.
Hence, it follows that the learned joint representation indeed contains rich information from other modalities.

\section{Conclusion and Future Work}
\label{sec:cncl}
In this paper, we propose two adaptive fusion techniques that allow for effective multimodal fusion.
Instead of ``fixing" a fusion operation a priori, we let the model decide ``how" to extract and effectively combine signals from different modalities.
Moreover, the joint multimodal representations learned by such models are empirically shown to be robust, which allows the system to maintain good performance even in the absence of some information.
Our results indicate that such adaptive models deliver without compromising performance than their more massive counterparts, such as transformers, which is a significant gain.

Our experiments indicate the importance of learning richer unimodal representations.
This also suggests that using these methods in conjunction with transformers, which may learn richer unimodal representations, should further improve downstream tasks such as multimodal machine translation and speech emotion recognition.
Currently, the attention mechanism is applied only on text.
So, another simple way to improve performance would be to introduce visual and acoustic attention mechanisms as well;
however, we would still need to address the core problem heterogeneity.

\textbf{On training GANs:}
Training GANs is known to be difficult.
However, we employed various tricks to ensure the training stability of models, especially the one's employing GAN-Fusion.
In fact, a musing line of exploration towards learning a better adaptive model could be to probe the implicit assumptions of GANs themselves.
GANs are known to exhibit numerous issues in practice \citep{arora2017generalization, sinn2018non}.
In \newcite{li2018implicit}, the authors argue the need to return to the principle of maximum likelihood, insisting on full recall.

It is essential to note that much is unknown about these models.
More concrete and sound reasoning for the success of these models will rely on two vital components:
1) understanding of the dynamics of the learned latent space, and 2) aligning multimodal features to address heterogeneity.
Both these components require more interpretable representations of the otherwise black-box models.

\section{Acknowledgments}
We acknowledge \href{https://www.computecanada.ca/home/}{Compute Canada's} GPU support for our experiments.
We also thank Dhruv Kumar for their meaningful suggestions on the manuscript.
Finally, we thank the anonymous reviewers for their constructive comments on this work.

\bibliography{anthology,eacl2021}

\begin{thebibliography}{39}
\expandafter\ifx\csname natexlab\endcsname\relax\def\natexlab#1{#1}\fi

\bibitem[{Arora et~al.(2017)Arora, Ge, Liang, Ma, and
  Zhang}]{arora2017generalization}
Sanjeev Arora, Rong Ge, Yingyu Liang, Tengyu Ma, and Yi~Zhang. 2017.
\newblock Generalization and equilibrium in generative adversarial nets (gans).
\newblock In \emph{ICML}, pages 224--232. PMLR.

\bibitem[{Banerjee and Lavie(2005)}]{banerjee2005meteor}
Satanjeev Banerjee and Alon Lavie. 2005.
\newblock Meteor: An automatic metric for mt evaluation with improved
  correlation with human judgments.
\newblock In \emph{ACL workshop on intrinsic and extrinsic evaluation measures
  for machine translation and/or summarization}, pages 65--72.

\bibitem[{Busso et~al.(2008)Busso, Bulut, Lee, Kazemzadeh, Mower, Kim, Chang,
  Lee, and Narayanan}]{busso2008iemocap}
Carlos Busso, Murtaza Bulut, Chi-Chun Lee, Abe Kazemzadeh, Emily Mower, Samuel
  Kim, Jeannette~N Chang, Sungbok Lee, and Shrikanth~S Narayanan. 2008.
\newblock Iemocap: Interactive emotional dyadic motion capture database.
\newblock \emph{Language resources and evaluation}, page 335.

\bibitem[{Chandar et~al.(2016)Chandar, Khapra, Larochelle, and
  Ravindran}]{chandar2016correlational}
Sarath Chandar, Mitesh~M Khapra, Hugo Larochelle, and Balaraman Ravindran.
  2016.
\newblock Correlational neural networks.
\newblock \emph{Neural computation}, 28(2):257--285.

\bibitem[{Cortes and Vapnik(1995)}]{cortes1995support}
Corinna Cortes and Vladimir Vapnik. 1995.
\newblock Support-vector networks.
\newblock \emph{Machine learning}, 20(3):273--297.

\bibitem[{{Elliott} et~al.(2016){Elliott}, {Frank}, {Sima'an}, and
  {Specia}}]{elliott-EtAl:2016:VL16}
D.~{Elliott}, S.~{Frank}, K.~{Sima'an}, and L.~{Specia}. 2016.
\newblock Multi30k: Multilingual english-german image descriptions.
\newblock In \emph{Proceedings of the 5th Workshop on Vision and Language},
  pages 70--74.

\bibitem[{Elliott et~al.(2017)Elliott, Frank, Barrault, Bougares, and
  Specia}]{ElliottFrankBarraultBougaresSpecia2017}
Desmond Elliott, Stella Frank, Lo\"{i}c Barrault, Fethi Bougares, and Lucia
  Specia. 2017.
\newblock {Findings of the Second Shared Task on Multimodal Machine Translation
  and Multilingual Image Description}.
\newblock In \emph{Proceedings of the Second Conference on Machine
  Translation}.

\bibitem[{Gao et~al.(2019)Gao, Lee, Zhang, Brockett, Galley, Gao, and
  Dolan}]{gao2019jointly}
Xiang Gao, Sungjin Lee, Yizhe Zhang, Chris Brockett, Michel Galley, Jianfeng
  Gao, and Bill Dolan. 2019.
\newblock \href
  {https://www.microsoft.com/en-us/research/publication/jointly-optimizing-diversity-and-relevance-in-neural-response-generation-2/}
  {Jointly optimizing diversity and relevance in neural response generation}.
\newblock In \emph{NAACL-HLT}.

\bibitem[{Goodfellow(2016)}]{goodfellow2016nips}
Ian Goodfellow. 2016.
\newblock \href {http://arxiv.org/abs/1701.00160} {{NIPS 2016 Tutorial}:
  Generative adversarial networks}.

\bibitem[{Gr{\"o}nroos et~al.(2018)Gr{\"o}nroos, Huet, Kurimo, Laaksonen,
  Merialdo, Pham, Sj{\"o}berg, Sulubacak, Tiedemann, Troncy, and
  V{\'a}zquez}]{gronroos2018memad}
Stig-Arne Gr{\"o}nroos, Benoit Huet, Mikko Kurimo, Jorma Laaksonen, Bernard
  Merialdo, Phu Pham, Mats Sj{\"o}berg, Umut Sulubacak, J{\"o}rg Tiedemann,
  Raphael Troncy, and Ra{\'u}l V{\'a}zquez. 2018.
\newblock \href {https://doi.org/10.18653/v1/W18-6439} {The {M}e{MAD}
  submission to the {WMT}18 multimodal translation task}.
\newblock In \emph{Proceedings of the Third Conference on Machine Translation:
  Shared Task Papers}, pages 603--611.

\bibitem[{Hochreiter and Schmidhuber(1997)}]{hochreiter1997long}
Sepp Hochreiter and J{\"u}rgen Schmidhuber. 1997.
\newblock Long short-term memory.
\newblock \emph{Neural computation}, 9(8):1735--1780.

\bibitem[{Huang et~al.(2018)Huang, Cho, Zhang, Ji, and Knight}]{huang2018multi}
Lifu Huang, Kyunghyun Cho, Boliang Zhang, Heng Ji, and Kevin Knight. 2018.
\newblock \href {https://doi.org/10.18653/v1/D18-1023} {Multi-lingual common
  semantic space construction via cluster-consistent word embedding}.
\newblock In \emph{EMNLP}, pages 250--260.

\bibitem[{Kingma and Ba(2015)}]{kingma2014adam}
Diederik~P. Kingma and Jimmy Ba. 2015.
\newblock \href {http://arxiv.org/abs/1412.6980} {Adam: {A} method for
  stochastic optimization}.
\newblock In \emph{ICLR}.

\bibitem[{Li and Malik(2018)}]{li2018implicit}
Ke~Li and Jitendra Malik. 2018.
\newblock On the implicit assumptions of gans.
\newblock \emph{NeurIPS Workshop on Critiquing and Correcting Trends in Machine
  Learning}.

\bibitem[{Liang et~al.(2019)Liang, Liu, Tsai, Zhao, Salakhutdinov, and
  Morency}]{liang2019learning}
Paul~Pu Liang, Zhun Liu, Yao-Hung~Hubert Tsai, Qibin Zhao, Ruslan
  Salakhutdinov, and Louis-Philippe Morency. 2019.
\newblock \href {https://doi.org/10.18653/v1/P19-1152} {Learning
  representations from imperfect time series data via tensor rank
  regularization}.
\newblock In \emph{ACL}, pages 1569--1576.

\bibitem[{Liu et~al.(2018)Liu, Shen, Lakshminarasimhan, Liang, Bagher~Zadeh,
  and Morency}]{liu2018efficientlow}
Zhun Liu, Ying Shen, Varun~Bharadhwaj Lakshminarasimhan, Paul~Pu Liang, AmirAli
  Bagher~Zadeh, and Louis-Philippe Morency. 2018.
\newblock \href {https://doi.org/10.18653/v1/P18-1209} {Efficient low-rank
  multimodal fusion with modality-specific factors}.
\newblock In \emph{ACL}, pages 2247--2256.

\bibitem[{Luong et~al.(2015)Luong, Pham, and Manning}]{luong2015effective}
Thang Luong, Hieu Pham, and Christopher~D. Manning. 2015.
\newblock \href {https://doi.org/10.18653/v1/D15-1166} {Effective approaches to
  attention-based neural machine translation}.
\newblock In \emph{EMNLP}, pages 1412--1421.

\bibitem[{Maas et~al.(2013)Maas, Hannun, and Ng}]{Maas13}
Andrew~L. Maas, Awni~Y. Hannun, and Andrew~Y. Ng. 2013.
\newblock Rectifier nonlinearities improve neural network acoustic models.
\newblock In \emph{ICML Workshop on Deep Learning for Audio, Speech and
  Language Processing}.

\bibitem[{Matthews et~al.(2002)Matthews, Cootes, Bangham, Cox, and
  Harvey}]{avletters}
Iain Matthews, Timothy~F Cootes, J~Andrew Bangham, Stephen Cox, and Richard
  Harvey. 2002.
\newblock Extraction of visual features for lipreading.
\newblock \emph{IEEE Transactions on Pattern Analysis and Machine
  Intelligence}, 24(2):198--213.

\bibitem[{Morade and Patnaik(2015)}]{Morade2015}
Sunil~S. Morade and Suprava Patnaik. 2015.
\newblock \href {https://doi.org/10.1016/j.ijleo.2015.08.192} {Comparison of
  classifiers for lip reading with cuave and tulips database}.
\newblock \emph{Optik - International Journal for Light and Electron Optics},
  126(24):5753--5761.

\bibitem[{Ngiam et~al.(2011)Ngiam, Khosla, Kim, Nam, Lee, and
  Ng}]{ngiam2011multimodal}
Jiquan Ngiam, Aditya Khosla, Mingyu Kim, Juhan Nam, Honglak Lee, and Andrew~Y
  Ng. 2011.
\newblock Multimodal deep learning.
\newblock In \emph{ICML}, pages 689--696.

\bibitem[{Papineni et~al.(2002)Papineni, Roukos, Ward, and
  Zhu}]{Papineni2002bleu}
Kishore Papineni, Salim Roukos, Todd Ward, and Wei-Jing Zhu. 2002.
\newblock \href {https://doi.org/10.3115/1073083.1073135} {Bleu: A method for
  automatic evaluation of machine translation}.
\newblock In \emph{ACL}, pages 311--318.

\bibitem[{Paszke et~al.(2019)Paszke, Gross, Massa, Lerer, Bradbury, Chanan,
  Killeen, Lin, Gimelshein, Antiga, Desmaison, Kopf, Yang, DeVito, Raison,
  Tejani, Chilamkurthy, Steiner, Fang, Bai, and Chintala}]{pytorch}
Adam Paszke, Sam Gross, Francisco Massa, Adam Lerer, James Bradbury, Gregory
  Chanan, Trevor Killeen, Zeming Lin, Natalia Gimelshein, Luca Antiga, Alban
  Desmaison, Andreas Kopf, Edward Yang, Zachary DeVito, Martin Raison, Alykhan
  Tejani, Sasank Chilamkurthy, Benoit Steiner, Lu~Fang, Junjie Bai, and Soumith
  Chintala. 2019.
\newblock Pytorch: An imperative style, high-performance deep learning library.
\newblock In \emph{NeurIPS}, pages 8024--8035.

\bibitem[{Patterson et~al.(2002)Patterson, Gurbuz, Tufekci, and
  Gowdy}]{patterson2002cuave}
Eric~K Patterson, Sabri Gurbuz, Zekeriya Tufekci, and John~N Gowdy. 2002.
\newblock Cuave: A new audio-visual database for multimodal human-computer
  interface research.
\newblock In \emph{ICASSP}, volume~2, pages II--2017.

\bibitem[{Pham et~al.(2019)Pham, Liang, Manzini, Morency, and
  P{\'o}czos}]{pham2019found}
Hai Pham, Paul~Pu Liang, Thomas Manzini, Louis-Philippe Morency, and
  Barnab{\'a}s P{\'o}czos. 2019.
\newblock Found in translation: Learning robust joint representations by cyclic
  translations between modalities.
\newblock In \emph{AAAI}, volume~33, pages 6892--6899.

\bibitem[{Povey et~al.(2011)Povey, Ghoshal, Boulianne, Burget, Glembek, Goel,
  Hannemann, Motlicek, Qian, Schwarz et~al.}]{povey2011kaldi}
Daniel Povey, Arnab Ghoshal, Gilles Boulianne, Lukas Burget, Ondrej Glembek,
  Nagendra Goel, Mirko Hannemann, Petr Motlicek, Yanmin Qian, Petr Schwarz,
  et~al. 2011.
\newblock The kaldi speech recognition toolkit.
\newblock In \emph{IEEE workshop on automatic speech recognition and
  understanding}.

\bibitem[{Salakhutdinov and Hinton(2009)}]{salakhutdinov2009deep}
Ruslan Salakhutdinov and Geoffrey Hinton. 2009.
\newblock Deep boltzmann machines.
\newblock In \emph{Artificial intelligence and statistics}, pages 448--455.

\bibitem[{Sanabria et~al.(2018)Sanabria, Caglayan, Palaskar, Elliott, Barrault,
  Specia, and Metze}]{sanabria2018how2}
Ramon Sanabria, Ozan Caglayan, Shruti Palaskar, Desmond Elliott, Lo\"ic
  Barrault, Lucia Specia, and Florian Metze. 2018.
\newblock \href {http://arxiv.org/abs/1811.00347} {{How2:} a large-scale
  dataset for multimodal language understanding}.
\newblock In \emph{Workshop on Visually Grounded Interaction and Language
  (ViGIL)}. NeurIPS.

\bibitem[{Sanh et~al.(2019)Sanh, Debut, Chaumond, and
  Wolf}]{sanh2019distilbert}
Victor Sanh, Lysandre Debut, Julien Chaumond, and Thomas Wolf. 2019.
\newblock Distilbert, a distilled version of bert: smaller, faster, cheaper and
  lighter.
\newblock \emph{arXiv preprint arXiv:1910.01108}.

\bibitem[{Shi et~al.(2019)Shi, Siddharth, Paige, and Torr}]{shi2019variational}
Yuge Shi, N~Siddharth, Brooks Paige, and Philip Torr. 2019.
\newblock Variational mixture-of-experts autoencoders for multi-modal deep
  generative models.
\newblock In \emph{NeurIPS}, pages 15718--15729.

\bibitem[{Simonyan and Zisserman(2015)}]{simonyan2014very}
Karen Simonyan and Andrew Zisserman. 2015.
\newblock Very deep convolutional networks for large-scale image recognition.
\newblock In \emph{ICLR}.

\bibitem[{Sinn and Rawat(2018)}]{sinn2018non}
Mathieu Sinn and Ambrish Rawat. 2018.
\newblock Non-parametric estimation of jensen-shannon divergence in generative
  adversarial network training.
\newblock In \emph{AISTATS}, pages 642--651.

\bibitem[{Srivastava and Salakhutdinov(2012)}]{srivastava2012multimodal}
Nitish Srivastava and Ruslan~R Salakhutdinov. 2012.
\newblock Multimodal learning with deep boltzmann machines.
\newblock In \emph{NeurIPS}, pages 2222--2230.

\bibitem[{Sutskever et~al.(2014)Sutskever, Vinyals, and
  Le}]{sutskever2014sequence}
Ilya Sutskever, Oriol Vinyals, and Quoc~V Le. 2014.
\newblock Sequence to sequence learning with neural networks.
\newblock In \emph{NeurIPS}, pages 3104--3112.

\bibitem[{Tsai et~al.(2019)Tsai, Bai, Liang, Kolter, Morency, and
  Salakhutdinov}]{tsai2019multimodal}
Yao-Hung~Hubert Tsai, Shaojie Bai, Paul~Pu Liang, J~Zico Kolter, Louis-Philippe
  Morency, and Ruslan Salakhutdinov. 2019.
\newblock Multimodal transformer for unaligned multimodal language sequences.
\newblock In \emph{ACL}, volume 2019, page 6558.

\bibitem[{Vaswani et~al.(2017)Vaswani, Shazeer, Parmar, Uszkoreit, Jones,
  Gomez, Kaiser, and Polosukhin}]{vaswani2017attention}
Ashish Vaswani, Noam Shazeer, Niki Parmar, Jakob Uszkoreit, Llion Jones,
  Aidan~N Gomez, {\L}ukasz Kaiser, and Illia Polosukhin. 2017.
\newblock Attention is all you need.
\newblock In \emph{NeurIPS}, pages 5998--6008.

\bibitem[{Yoon et~al.(2019)Yoon, Byun, Dey, and Jung}]{yoon2019speech}
Seunghyun Yoon, Seokhyun Byun, Subhadeep Dey, and Kyomin Jung. 2019.
\newblock Speech emotion recognition using multi-hop attention mechanism.
\newblock In \emph{ICASSP}, pages 2822--2826.

\bibitem[{Yoon et~al.(2018)Yoon, Byun, and Jung}]{yoon2018multimodal}
Seunghyun Yoon, Seokhyun Byun, and Kyomin Jung. 2018.
\newblock Multimodal speech emotion recognition using audio and text.
\newblock In \emph{SLT Workshop}, pages 112--118. IEEE.

\bibitem[{Zadeh et~al.(2017)Zadeh, Chen, Poria, Cambria, and
  Morency}]{zadeh2017tensor}
Amir Zadeh, Minghai Chen, Soujanya Poria, Erik Cambria, and Louis-Philippe
  Morency. 2017.
\newblock \href {https://doi.org/10.18653/v1/D17-1115} {Tensor fusion network
  for multimodal sentiment analysis}.
\newblock In \emph{EMNLP}, pages 1103--1114.

\end{thebibliography}
\bibliographystyle{acl_natbib}

\appendix

\end{document}